\documentclass[conference]{IEEEtran}
\usepackage{times}

\usepackage[numbers]{natbib}
\usepackage{multicol}
\usepackage[bookmarks=true]{hyperref}

\usepackage{graphicx}
\usepackage{url}
\usepackage{subfigure}

\usepackage{algorithm}
\usepackage[noend]{algpseudocode}

\usepackage{tikz}
\usetikzlibrary{arrows,automata}
\usepackage{array,multirow,graphicx}

\newcommand{\methodName}[0]{RAQ}

\usepackage{amsmath}
\DeclareMathOperator*{\argmax}{arg\,max}

\begin{document}

\title{End-to-End Robotic Reinforcement Learning \\ without Reward Engineering}

\author{\authorblockN{Avi Singh, Larry Yang, Kristian Hartikainen, Chelsea Finn, Sergey Levine}
\authorblockA{University of California, Berkeley\\
Email: \{avisingh, larrywyang, hartikainen, cbfinn, svlevine\}@berkeley.edu}
}

\maketitle

\IEEEpeerreviewmaketitle

\begin{abstract}
The combination of deep neural network models and reinforcement learning algorithms can make it possible to learn policies for robotic behaviors that directly read in raw sensory inputs, such as camera images, effectively subsuming both estimation and control into one model.
However, real-world applications of reinforcement learning must specify the goal of the task by means of a manually programmed reward function, which in practice requires either designing the very same perception pipeline that end-to-end reinforcement learning promises to avoid, or else instrumenting the environment with additional sensors to determine if the task has been performed successfully. In this paper, we propose an approach for removing the need for manual engineering of reward specifications by enabling a robot to learn from a modest number of examples of successful outcomes, followed by actively solicited queries, where the robot shows the user a state and asks for a label to determine whether that state represents successful completion of the task. While requesting labels for every single state would amount to asking the user to manually provide the reward signal, our method requires labels for only a tiny fraction of the states seen during training, making it an efficient and practical approach for learning skills without manually engineered rewards. We evaluate our method on real-world robotic manipulation tasks where the observations consist of images viewed by the robot's camera. In our experiments, our method effectively learns to arrange objects, place books, and drape cloth, directly from images and without any manually specified reward functions, and with only 1-4 hours of interaction with the real world. Videos of learned behavior are available at \href{https://sites.google.com/view/reward-learning-rl/}{sites.google.com/view/reward-learning-rl/}. 
\end{abstract}

\section{Introduction}

Reinforcement learning holds the promise of enabling robotic systems to improve continuously through experience. A robot equipped with a sufficiently powerful reinforcement learning algorithm can become more and more proficient at a given task as it practices it directly in the real world. With end-to-end algorithms that enable skill learning directly from raw sensory observations, such as images, this capability can be tremendously powerful: every aspect of the robot's estimation and control pipeline can improve from experience and become better and better suited to the task at hand. However, in order to enable such continual improvement, the robot must have a way to evaluate whether it is succeeding at the task -- it needs a reward function. Unfortunately, in practice, the design of reward functions for robotic skills is very challenging, especially when learning skills from raw observations such as images: manually defining reward functions typically requires manually-designed perception systems or instrumentation of the environment (e.g., by placing additional sensors)~\cite{progressive, yahya2017collective, pancake, schenck2017visual}. This either precludes learning directly in open-world environments that are not instrumented, or makes learning entirely contingent on the reward perception system, which does not improve or adapt over the robot's lifetime. In many scenarios, the need for manual engineering of reward functions defeats the point of end-to-end learning from pixels, if the reward function itself requires a prior perception pipeline or instrumentation.

\begin{figure}
\begin{center}
    \includegraphics[width=\linewidth]{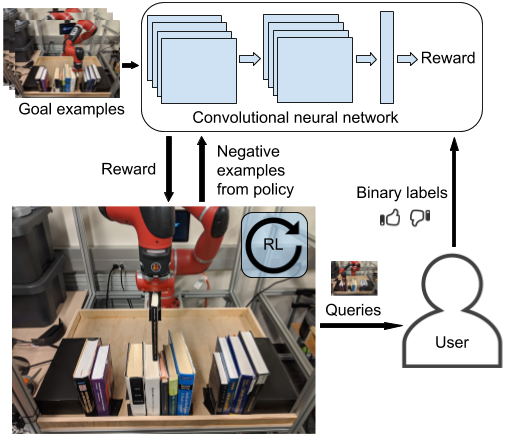}
    \caption{\textbf{Illustration of our approach.} We learn a reward function on high-dimensional observations (i.e. pixels) using a neural network classifier. The classifier receives a set of goal examples (images) from a user that specify the desired outcome. Negative examples for classifier training are obtained from the policy. Our method trains a policy to place a book on the bookshelf via reinforcement learning w.r.t. the learned reward. During this process, the robot periodically queries the user with images, and the user provides a binary label indicating whether or not this image corresponds to a successful outcome. This allows the robot to learn tasks with a modest number of examples and queries, and without manually designing reward functions.}
    \label{fig:teaser}
\end{center}
\end{figure}

Inverse reinforcement learning methods~\cite{Ziebart08, Wulfmeier15, gcl, gail, airl} seek to automate reward definition by learning a reward function from expert demonstrations. However, these methods have limitations of their own: collecting expert demonstrations puts a significant data collection burden on the user, and demonstrations are usually non-intuitive for a person to provide (often involving kinesthetically moving a robot~\cite{gcl} or teleoperating it~\cite{Vasquez14}). More importantly, demonstrating how a task is done defeats a central goal of reinforcement learning: autonomously discovering how to perform skills through trial and error. An alternative way to specify goals is to provide images of the goal, and then train a goal classifier on this data~\cite{Vecerik18, flo}. The success probabilities from this classifier can then be used as a reward for training reinforcement learning agents to achieve the specified goal. While this is appealing in principle, a na\"{i}vely-trained classifier can easily be exploited by a reinforcement learning algorithm: the RL algorithm can visit parts of the observation space that the classifier was not trained on, causing it to output incorrect probabilities~\cite{vice}. 

A recent approach, variational inverse control with events (VICE)~\cite{vice}, mitigates the exploitation issues of na\"{i}ve classifiers by adversarially mining negative examples from the policy's own experience, analogously to adversarial imitation learning techniques~\cite{gcl,gail,airl}. However, VICE relies entirely on the positive outcome examples provided at the beginning of training to understand the task, which in practice means that a large number of examples is needed. Further, because VICE relies on on-policy RL for training the policy and for gathering negative examples for the classifier, it requires millions of samples, which may be impractical in the real world. 
In this paper, we address both of these issues to enable end-to-end reinforcement learning on real robots from pixel observations, and without any task-specific engineering for obtaining rewards. To remove the reliance on a large amount of positive examples provided up front, we elicit a small number of additional queries from a human user as the robot collects additional experience. These active queries are selected based on uncertainty estimates from the classifier that is being used as a reward function, and allow us to learn effective rewards from a small number of initial examples. Further, we extend both the policy learning and the classifier training procedure in VICE to the off-policy setting, which allows us to learn robotic skills in the real world with only 1-4 hours of interaction time, entirely from image observations.

The primary contribution of this paper is a framework for learning robotic skills from high-dimensional observations, such as images, without hand-designing reward functions.
Our method uses a small number of examples of positive outcomes (without demonstrations), followed by a modest number of additional binary active queries, where the robot asks the user if a particular outcome is successful or not. Our approach is based on efficient off-policy reinforcement learning, making it well-suited for real-world learning. 
Our experiments demonstrate that our method can learn a variety of real-world robotic manipulation skills, directly from images, and directly in the real world. Results include draping cloth over an object, placing books on a bookshelf, and pushing mugs onto a coaster. Learning requires minimal user supervision and only 1-4 hours of interaction time, which is substantially less than that of prior work~\cite{haarnoja2018sacapps, qt-opt, pinto2016supersizing, Levine16googlegrasping, Ebert18journal}.
\section{Related Work}

Reinforcement learning has been applied to a wide variety of robotic manipulation tasks, including grasping objects~\cite{qt-opt}, in-hand object manipulation~\cite{dexterity, VanHoofInHandManipulation, dexterous1, dexterous2},
manipulating fluids~\cite{schenck2017visual}, door opening~\cite{yahya2017collective, dooropening},
and cloth folding~\cite{matas18}.
However, applications of RL in the real world require considerable effort to design and evaluate the reward function. For example, using thermal cameras for tracking fluids~\cite{schenck2017visual}, mocap sensors~\cite{pancake} or computer vision systems~\cite{progressive} for tracking objects, and accelerometers for determining the state of a door~\cite{yahya2017collective}. Since such instrumentation needs to be done for any new task that we may wish to learn, it poses a significant bottleneck to widespread adoption of reinforcement learning for robotics, and precludes the use of these methods directly in open-world environments that lack this instrumentation.

Data-driven approaches for reward specification~\cite{ng_and_russel,ng_and_abbeel,gcl, gail, airl, ziebart, ratliff_bagnell,edwards2017perceptual,Majumdar-RSS-17, Christiano17, Jain13}
seek to overcome this issue, but typically require demonstration data to acquire rewards. Such data can be onerous and time-consuming for users to provide. Recent work on active learning for inverse RL has sought to reduce the required number of demonstrations~\cite{Brown18, Cui18, Lopes09, Cohn11}, but still requires some number of demonstrations to be provided manually. Our method only requires a modest number of examples of successful outcomes, followed by binary queries where the user indicates whether a particular outcome that the robot achieved is successful or not. Both of these can be provided easily, without any teleoperation or kinesthetic teaching. 
Related to our method, \citet{Daniel-RSS-14} propose to learn rewards from active queries that elicit numerical scores. In contrast to this approach, our method uses only binary success queries, which are easier to provide, and can be readily combined with deep networks for learning skills from image observations. Another line of research that queries binary feedback from humans is that of learning from human preferences~\cite{Jain13, Christiano17}, but these techniques have so far proven to be quite expensive in the deep reinforcement learning setting in terms of both the supervision needed from humans, and the overall sample complexity. Even in simulation with low-dimensional observations, \citet{Christiano17} make about one thousand queries from humans (we make 50-75 such queries in the real world), and require tens of millions of timesteps of interaction with the environment (we require around tens of thousands of such interactions). Furthermore, comparing trajectories is often a harder form of feedback to elicit from humans, especially towards the start of training where all trajectories might be equally undesirable. 

Classifier training serves as an alternative to inverse reinforcement learning for data-driven reward specification~\cite{Vecerik18, flo,pinto2016supersizing}. However, using classifier-based rewards for reinforcement learning is prone to exploitation by RL agents, as such agents quickly capitalize on any imperfections in the learned classifier~\cite{vice, flo}. VICE~\cite{vice} overcomes this issue issue by adversarially mining negatives form the learned policy, but usually requires a large number of samples to learn due to using an on-policy algorithm~\cite{Schulman15} for policy improvement and classifier updates. Furthermore, it usually utilizes a large number of goal examples (on the order of 50K for image observations) for successfully learning the task. Our approach overcomes both of these issues, and we demonstrate that it can be used for practical real-world reinforcement learning.

\newcommand{\data}{\mathcal{D}}
\newcommand{\obs}{\mathbf{s}}
\newcommand{\state}{\mathbf{s}}
\newcommand{\act}{\mathbf{a}}
\newcommand{\posdata}{\data^+}
\newcommand{\loss}{\mathcal{L}}
\newcommand{\out}{y}
\newcommand{\policy}{\pi}
\newcommand{\E}{\mathbb{E}}

\section{Preliminaries}
Our aim is to make it possible to specify reward functions for robotic reinforcement learning with a small number of outcome examples (e.g., photographs of a successful task outcome in the case of image-based RL), followed by a modest number of \emph{active queries}, where the robot asks the user if a particular outcome is a success or not. Here, we summarize the framework of robotic reinforcement learning and introduce how classifiers can be used as reward functions. In RL, the goal is to learn a policy in a Markov decision process, which in a robotic control problem consists of a state space $\mathcal{S}$ and an action space $\mathcal{A}$. In our case, the state space consists of image observations, and the action space consists of desired end-effector motion. The policy, denoted $\policy(\act_t | \state_t)$, chooses actions, and the state is assumed to evolve according to the unknown dynamics $p(\state_{t+1}|\state_t,\act_t)$. This generates a trajectory of states and actions $\tau: (\state_0, \act_0, \state_1, \act_1, ...)$. The goal in reinforcement learning is to optimize the expected total reward of the trajectory distribution induced by the policy.

\paragraph{Maximum Entropy RL} The particular reinforcement learning framework that we will use, called maximum entropy RL~\cite{Ziebart10,Haarnoja2017,Haarnoja2018}, also maximizes the entropy of the resulting distribution, resulting in the modified objective
\begin{equation}
    J(\policy) = \sum_{t = 0}^{T} E_{\tau\sim \policy}\left[{r(\obs_t, \act_t) - \log \policy(\act_t|\obs_t)}\right].
    \label{eq:max_ent_objective}
\end{equation}
While our framework could be combined with standard RL as well, the maximum entropy framework offers two benefits: first, maximum entropy RL algorithms, such as the off-policy soft actor-critic (SAC) algorithm that we use in our experiments~\cite{Haarnoja2018}, tend to produce stable and robust policies for real-world reinforcement learning, and second, maximum entropy RL makes it straightforward to integrate our method with VICE~\cite{vice}, a prior approach for classifier-based rewards, which we discuss in Section~\ref{sec:vice_event_learning}. An outline of SAC is presented in Algorithm~\ref{alg:sac}. SAC uses a replay buffer $\mathcal{R}$ to store past transitions, and trains both a critic and a maximum entropy actor on batches sampled from this buffer, while at the same time collecting experience with the current stochastic policy, which is represented as a Gaussian with the mean given by a neural network function of the state.

\begin{algorithm}[H]
    \caption{Soft actor-critic (SAC)}
    \small
    \label{alg:sac}
    \begin{algorithmic}[1]
    \State Initialize policy $\policy$, critic $Q$
    \State Initialize replay buffer $\mathcal{R}$
    \For {each iteration} 
    \For {each environment step}
    \State $\act_t \sim \policy_{\theta} (\act_t | \obs_t)$
    \State $\obs_{t+1} \sim  p(\obs_{t+1} | \obs_t, \act_t)$
    \State $\mathcal{R} \leftarrow \mathcal{R} \cup \{\left(\obs_t, \act_t, r(\obs_t, \act_t), \obs_{t+1}\right)\}$
    \EndFor
    \For {each gradient step}
    \State Sample from $\mathcal{R}$
    \State Update $\policy$ and $Q$ according to~\citet{Haarnoja2018}
    \EndFor
    \EndFor
    \end{algorithmic}
\end{algorithm}

\paragraph{Classifier-Based Rewards} Engineering reward functions for RL algorithms is difficult, especially when using image observations, because it requires identifying the state of the objects in the world and formulating a reliable success condition programmatically. Indeed, it is often easier for users to state whether a given outcome is successful or not than to write a program that will do so automatically. However, current RL algorithms require so many episodes that labeling the reward in each one manually would be impractical. A reasonable alternative is to use a goal classifier~\cite{flo, Vecerik18}, where the user provides a dataset of example states (e.g., images) before training the policy, denoted $\data := \{(\state_n, \out_n)  \}$, and a binary classifier $g(\state)$ is trained to predict whether a given state is a success or failure. The classifier-based RL framework is then summarised in Algorithm~\ref{alg:classifierRL}. Here, $\out_n$ represents the binary success or failure label, and $\loss$ denotes a binary classification loss (for example, the cross-entropy loss). Once trained, the classifier can be used to provide a reward during reinforcement learning. If the classifier provides a distribution $p_g(\out | \state)$, then a particularly convenient form for the reward is given by $\log p_g(\out | \state)$. As discussed in prior work~\cite{vice}, this has an appealing theoretical interpretation based on a connection to control as inference~\cite{controlasinference}, and in practice can provide some degree of shaping, as log-probabilities often increase smoothly as the agent approaches the goal.

\begin{algorithm}[H]
    \caption{Classifier-based rewards for RL}
    \small
    \label{alg:classifierRL}
    \begin{algorithmic}[1]
    \Require: $\data_i := \{(\state_n, \out_n)  \}$
    \State Update the parameters of $g$ to minimize $\sum_n \loss(g(\state_n), \out_n)$
    \State Run RL or planning, using reward derived from $\log p_g(\out|\state)$
    \end{algorithmic}
\end{algorithm}

Prior work generally requires both successful and failed examples to be part of $\data$~\cite{flo,Vecerik18}. When a policy is trained with this classifier, the policy can learn to \emph{exploit} the classifier, reaching states that are different from those that the classifier was trained on and fool it into outputting a success label erroneously~\cite{vice}.
The degree of exploitation is strongly dependent on how the negative examples are provided, and can only be avoided if a comprehensive set of negative examples covering the entire state space is supplied. In the next section, we will describe an approach which does not require a comprehensive set of negative examples up front, and instead uses a modest number of active queries from the user to address the exploitation problem.

\section{Reinforcement Learning with Active Queries}
\label{sec:method}

The goal of our method, which we call reinforcement learning with active queries (RAQ), is to learn robotic skills via reinforcement learning without requiring hand-engineered reward functions, using data that can be easily obtained from the user. More specifically, we train classifiers to distinguish between goal and non-goal observations, and use them to compute rewards. Instead of learning this classifier from a pre-specified static dataset alone (as done in prior work~\cite{Vecerik18,flo}), we introduce an active learning framework that queries a user for binary success labels for states that it would like to obtain ground truth labels for. This addresses two major challenges with classifier-based rewards: it removes the need for the user to provide a comprehensive set of negative examples up front, and it mitigates the classifier exploitation problem. 
Let $\{\obs_n\}_{n=1}^{t}$ denote the set of states that the agent encounters over the learning process, where $t$ is the total number of environment steps that the agent has taken so far. At any given step $t$, our algorithm decides which states from $\{\obs_n\}_{n=1}^{t}$ (if any) it should query a label for. We first introduce and motivate our query mechanism, and we then show how it can be combined with a classifier-based reward learning technique to obtain a practical algorithm for reinforcement learning in the real world.

\subsection{Active Queries}
\label{sec:active_query}

If the robot requests user labels for every single state it sees, it will have a very accurate reward. However, a typical RL run will collect tens or even hundreds of thousands of states worth of data, as discussed in Section~\ref{sec:experiments_real}, and labeling all of these states is impractical. Minimizing the required number of queries depends critically on the mechanism that decides which state should be labeled. The active learning literature provides a few potential mechanisms based on uncertainty, such as the maximum entropy heuristic. In practice, we found that the maximum entropy heuristic does not actually produce very good results, since the goal is not so much to obtain accurate goal classification everywhere, but rather to eliminate the ``exploitation'' problem, such that the classifier does not output false positives. To that end, we found that the most effective mechanism to select which states to label was to select the previously-unlabeled states with the highest probability of success according to the classifier. Recall that our reward is provided by a binary classifier, which specifies a distribution $p_g(y|\state)$, where $y$ is a binary variable indicating success. Following \citet{vice}, the reward is given by $\log p_g(y|\state)$. We can select the state $\state_k$ to label from the set of observed states according to
$$ k = \argmax_t~{\log p_{g}(\out=1|\obs_t)}~\forall\, t \text{ since last query}.$$ 
For most practical tasks, this query mechanism is also much more selective than the maximum entropy rule. First, negative examples are much easier to obtain than positive examples for most tasks, so requesting labels for only the potential positive examples avoids superfluous queries for high-entropy states that are unlikely to be informative of success. Second, because the policy is explicitly trying to visit states with high $p_g(y|\state)$, classifier exploitation is due to false positives rather than negatives~\cite{vice}. Since only states with positive predictions can be false positives, querying for labels for these states is an effective mechanism for mitigating classifier exploitation.

Aside from selecting which states to label, we must also choose how often to request labels. We adopt a simple scheme where labels are queried at fixed intervals. We found that simply choosing a query frequency based on the expected training length and a query budget was sufficient. For the real-world robot experiments presented this paper, we adjust the frequency so that we make between 25 to 75 active queries for a single run of a reinforcement learning experiment. Details on this can be found in Section~\ref{sec:experiments_real}.

\subsection{Classifier-Based RL with Active Queries}
\label{sec:raq_description}
We now explain how we combine our active query framework with classifier-based rewards for reinforcement learning. Our approach is summarized in Algorithm~\ref{alg:activeClassifierRL}. Similar to standard classifier reward-based RL, we first train a classifier $g$ on an initial dataset $\data$.
The RL algorithm then uses $\log p_g(\out|\state)$ as the reward, and runs for a predefined number of time steps, at which point we select a new state to query by selecting the state with the largest value for $\log p_g(\out|\state)$. This labeled state is added to the dataset $\data$, and the classifier is then fine-tuned. We then continue training with RL, and repeat the process. This procedure is repeated until convergence or until a fixed budget of samples or queries is exceeded.

\begin{algorithm}[H]
    \caption{Reinforcement learning with active queries (RAQ)}
    \small
    \label{alg:activeClassifierRL}
    \begin{algorithmic}[1]
    \Require initial $\data := \{(\obs_n, \out_n)  \}$
    %\State Update the parameters of $g$ to minimize $\sum_{n} \loss(g(\obs_n), \out_n)$
    \State Initialize policy $\policy$, critic $Q$
    \State Initialize replay buffer $\mathcal{R}$
    %\State Initialize replay buffer $\mathcal{R}$
    \For {each iteration} 
    \For {each environment step}
    \State $\act_t \sim \pi_{\theta} (\act_t | \obs_t)$
    \State $\obs_{t+1} \sim  p(\obs_{t+1} | \obs_t, \act_t)$
    %\State $r(\obs_t) \leftarrow g(\obs_t) $
    \State $\mathcal{R} \leftarrow \mathcal{R} \cup \{\left(\obs_t, \act_t, \obs_{t+1}\right)\}$
    \EndFor
    \For {each gradient step}
    \State Sample from $\mathcal{R}$
    \State Compute rewards: $r(\obs_t) \leftarrow \log p_g(\out_t | \obs_t)$
    \State Update $\policy$ and $Q$ according to~\citet{Haarnoja2018}
    % \State Update $\pi, \psi, \hat{\psi}, \phi$ according to \citet{Haarnoja2018}
    \EndFor
    
    \If {active query}
    \State $k \rightarrow \argmax{\log p_g(\out_t | \obs_t)}$ for all $t$ since the last query
    \If {$\obs_k$ is a successful outcome}
        \State $\data \leftarrow \data \cup \{(\obs_k, 1)\}$
    \Else
        \State $\data \leftarrow \data \cup \{\left(\obs_k, 0\right)\}$
    \EndIf 

    \State Update $g$ to minimize $\sum_{n} \loss(g(\obs_n), \out_n) $ %\forall \obs_n \in \data
    \EndIf
    \EndFor
    \end{algorithmic}
\end{algorithm}

\section{Off-Policy VICE with Active Queries}
\label{sec:method_vice}

While standard RAQ, as described in Section~\ref{sec:method}, is effective in mitigating the classifier exploitation problem and can enable reinforcement learning with classifier-based rewards, it only utilizes a very small fraction of the data that is collected when running RL. RL algorithms typically collect tens or hundreds of thousands of transitions when learning to solve a robotic task in the real world, but \methodName{} only makes tens of queries, barely using 0.1\% of the data collected. Ideally, we would like to make use of \emph{all} the data collected during RL. In this section, we first review the basic VICE algorithm, which was proposed by~\citet{vice}. VICE is classifier-based reward specification framework that uses on-policy RL with policy gradients, and generally requires a large number of positive outcome examples. However, VICE can effectively overcome the classifier exploitation problem, and does so by using all of the data collected during RL without making any active queries. We first show how VICE can be extended into the off-policy setting, providing a practical method for robotic RL. We then discuss how \methodName{} can be combined with VICE. The resulting method, which we call VICE-\methodName{}, combines the best properties of both techniques.

\subsection{VICE}
\label{sec:vice_event_learning}

%\begin{wrapfigure}{r}{0.5\textwidth}
\begin{figure}

\vspace{-0.05in}

\centering
\begin{tikzpicture}[->,auto,node distance=1.3cm,]
\node[state](S1){$\obs_1$}; %% CF: s_1:3, a_1:3 were not bolded before
\node[state](S2)[right of=S1]{$\obs_2$};
\node[state](S3)[right of=S2]{$\obs_3$};
\node[state](A1)[below of=S1]{$\act_1$};
\node[state](A2)[below of=S2]{$\act_2$};
\node[state](A3)[below of=S3]{$\act_3$};
\node[state](E1)[below of=A1,fill=gray!30]{$y_1$};
\node[state](E2)[below of=A2,fill=gray!30]{$y_2$};
\node[state](E3)[below of=A3,fill=gray!30]{$y_3$};
\node(DOTS)[right of=S3]{\ldots};
\node(DOTA)[right of=A3]{\ldots};
\node(DOTE)[right of=E3]{\ldots};
\node[state](ST)[right of=DOTS]{$\obs_t$};
\node[state](AT)[below of=ST]{$\act_t$};
\node[state](ET)[below of=AT,fill=gray!30]{$y_T$};

\path (S1) edge node {} (S2)
           edge node {} (A1)
           edge [bend right] node {} (E1)
      (A1) edge node {} (S2)
           edge node {} (E1)
      (S2) edge node {} (S3)
           edge node {} (A2)
           edge [bend right] node {} (E2)
      (A2) edge node {} (S3)
           edge node {} (E2)
      (S3) edge node {} (A3)
           edge node {} (DOTS)
           edge [bend right] node {} (E3)
      (A3) edge node {} (E3)
           edge node {} (DOTS)
      (DOTS) edge node {} (ST)
      (DOTA) edge node {} (ST)
      (ST) edge node {} (AT)
           edge [bend right] node {} (ET)
      (AT) edge node {} (ET)
          ;
\end{tikzpicture}
\caption{ \small
\label{fig:controlgraph}
A graphical model framework for VICE. The node $y_t$ is a binary random variable that denotes whether an event happens at a given time step or not.}
%In maximum entropy reinforcement learning, we observe $e_{1:T}=1$ and can perform inference on the trajectory to obtain a policy.
\vspace{-0.15in}
\end{figure}
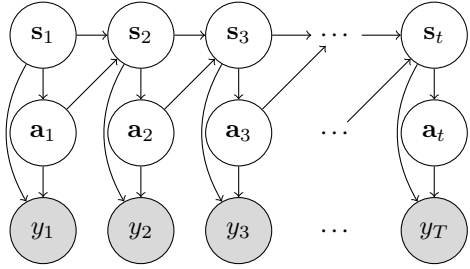

VICE~\cite{vice} is based on a formulation of reinforcement learning as inference in a graphical model, which is shown in Figure~\ref{fig:controlgraph}. In addition to the states and actions, this graphical model also includes binary event variables, $y_t$. Intuitively, these variables denote whether a particular event has taken place at time $t$, which in our case corresponds to successful completion of the task. We can formulate the problem of learning a policy that succeeds at the task as inference in this graphical model, where the policy corresponds to $p(\act_t|\obs_t,y_{1:T}=1)$. This follows from the framework of maximum entropy reinforcement learning~\cite{Ziebart08,controlasinference,vice}, and corresponds exactly to the maximum entropy objective in Equation~\ref{eq:max_ent_objective} with the reward given by $\log p(y_t=1 | \obs_t,\act_t)$, which is also the reward used by \methodName{}.

Learning the event probabilities in VICE corresponds to an optimization that is similar to maximum entropy inverse reinforcement learning~\cite{Ziebart10}. In the case of high-dimensional and continuous state spaces, as is the case for robotic reinforcement learning, a scalable way to implement maximum entropy inverse RL is to utilize adversarial inverse reinforcement learning (AIRL)~\cite{airl}. AIRL alternates between training a \emph{discriminator} to discriminate between the positive examples and the current policy's rollouts, and optimizing the policy with respect to the maximum entropy objective in Equation~\ref{eq:max_ent_objective}, using $\log p(y_t=1 | \obs_t,\act_t)$ as the reward. The discriminator in AIRL is parameterized by $\psi$ and given by the following equation:

\begin{equation}
D_\psi(\obs, \act) = \frac{\exp (f_\psi(\obs, \act)) }{ \exp (f_\psi(\obs, \act)) + \pi(\act|\obs)}.\label{eq:disc}
\end{equation}
As shown by Fu et al.~\cite{vice}, $f_\psi(\obs, \act)$ recovers $\log p(e=1|\obs,\act)$ at convergence of this adversarial learning procedure. The basic VICE algorithm is implemented as an on-policy reinforcement learning procedure, typically using policy gradient methods such as TRPO~\cite{Schulman15}. This makes it difficult to use for real-world robotic learning. Furthermore, VICE requires the success examples to include both the state $\obs$ and action $\act$, which is unnatural for a user to provide. We defer further details about the basic VICE algorithm to prior work due to space constraints~\cite{vice}. In the following sections, we describe our novel extension of VICE that lifts both of these limitations, and then present our complete VICE-\methodName{} algorithm, which combines VICE with active queries.

\subsection{Off-Policy VICE}
\label{sec:off-policy-vice}

In order to make VICE practical to use for real-world robotic learning, we must extend it so as to make it compatible with efficient off-policy deep reinforcement learning methods, and remove the need to obtain ground truth action labels for the positive examples, so that the user can readily specify examples simply by showing the robot example images of successful outcomes. Extending VICE to the off-policy setting first requires an off-policy reinforcement learning algorithm that can optimize the maximum entropy objective in Equation~\ref{eq:max_ent_objective}. The soft actor-critic algorithm~\cite{Haarnoja2018} provides one such method. Next, we need a way to train the discriminator. While in principle this would require importance sampling if using off-policy data from the replay buffer $\mathcal{R}$, prior work has observed that adversarial IRL can drop the importance weights both in theory~\cite{Finn16b} and in practice~\cite{Kostrikov18}. We adopt the same approach, and sample negative examples for the discriminator directly from $\mathcal{R}$ without importance weighting.

The standard VICE algorithm also requires the user-specified success examples to consist of state-action tuples $(\obs,\act)$, since both $\obs$ and $\act$ are required to update the discriminator $D_\theta(\obs,\act)$ when it has the form in Equation~\ref{eq:disc}. Even when $f_\psi(\obs)$ does not depend on the action, the term $\pi(\act|\obs)$ in the denominator does. Providing the actions is unnatural to the user, since the examples consist of isolated individual states showing successful outcomes (e.g., images of successful outcomes). Therefore, we remove the need for the user to specify actions by integrating out the actions for the positive examples in the VICE discriminator update, by using the current policy $\policy(\act|\obs)$. This amounts to sampling the actions for the positives, denoted $\act^E_i$, from $\policy(\act|\obs_i^E)$ as shown on line~\ref{alg:action_sampling} in Algorithm~\ref{alg:activeVICE}. 
At convergence, since $\pi(\act|\obs)$ approaches the expert's policy, this simplification produces the same action distribution, and therefore this update has the same fixed point as when the user supplies ground truth actions.

\subsection{Off-Policy VICE-\methodName{}}
\label{sec:viceraq_description}

Since the set of positives in VICE is fixed at the start of the algorithm, it typically requires a large set of positive examples provided by the user to begin with in order to prevent overfitting, sometimes as many as several thousand~\cite{vice}. By integrating VICE with our active query framework, we can substantially decrease the number of examples that are required, at the cost of several tens of binary queries, as in standard \methodName{}. To integrate \methodName{} with VICE, we simply add the active queries, as in Algorithm~\ref{alg:activeClassifierRL}, and append the labeled state to the example set if the label is positive. If the label is negative, there is no need to append the state, since VICE already uses all sampled states as negatives. The full VICE-\methodName{} algorithm is summarized in Algorithm~\ref{alg:activeVICE}. We start out with a dataset $\data$ consisting of positive examples.
Every iteration, we collect data from the environment and use it to update $f_\psi$, the policy, and the Q-function. At fixed intervals, we query the user using our active query mechanism discussed in Section~\ref{sec:active_query}, and update our dataset $\data$ if the queried state is labeled as a successful outcome. We continue running RL and updating the event probabilities $f_\psi$, utilizing both our initial dataset and any positives that we obtained from successful queries. This procedure continues until the policy converges, or after a specific period of time.

\begin{algorithm}[H]
    \caption{Off-Policy VICE-\methodName{} with soft actor-critic}
    \small
    \label{alg:activeVICE}
    \begin{algorithmic}[1]
    \Require: $\data_i := \{(\obs_n, 1)  \}$
    \State Initialize $f_\psi$ (described in Equation~\ref{eq:disc})
    \State Initialize policy $\policy$, critic $Q$
    \State Initialize replay buffer $\mathcal{R}$
    \For {each iteration} 
    \For {each environment step}
    \State $\act_t \sim \pi_{\theta} (\act_t | \obs_t)$
    \State $\obs_{t+1} \sim  p(\obs_{t+1} | \obs_t, \act_t)$
    \State $\mathcal{R} \leftarrow \mathcal{R} \cup \{\left(\obs_t, \act_t, \obs_{t+1}\right)\}$
    \EndFor
    \For {each gradient step for $f_\psi$}
    \State Sample positives from $\data$
    \State Sample action labels $\act_i^E \sim \policy(\act|\obs_i^E)$ \label{alg:action_sampling}
    \State Sample negatives from $\mathcal{R}$
    \State Update $f_\psi$ using Equation~\ref{eq:disc} as discriminator 
    \EndFor

    \For {each gradient step for the policy $\policy$}
    \State Sample from $\mathcal{R}$
    \State Compute rewards: $r(\obs_t) \leftarrow f_\psi(\obs_t)$
    \State Update $\policy$ and $Q$ according to~\citet{Haarnoja2018}
    
    \EndFor
    
    \If {active query}
    \State $k \rightarrow \argmax{f_\psi(\obs_t)}$ for all $t$ since the last query
    \If {$\obs_k$ is a successful outcome}
        \State $\data \leftarrow \data \cup \{(\obs_k, 1)\}$

    \EndIf 
    \EndIf 

    \EndFor
    \end{algorithmic}
\end{algorithm}

\section{VICE-RAQ for Image-based Manipulation}

\begin{figure}
\begin{center}
    \hspace{-0.1cm}
    \includegraphics[width=0.98\linewidth]{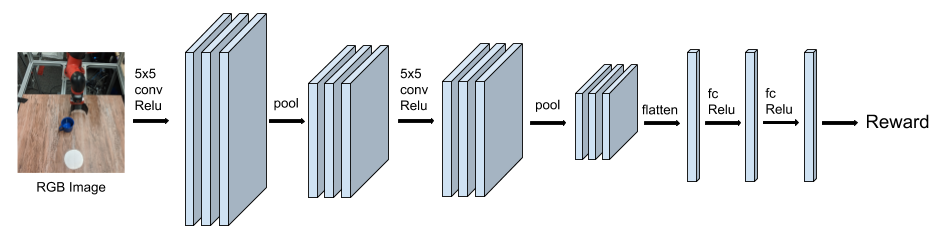}
    \caption{Our convolutional neural network architecture. The same architecture is used for the policy, critic, and the learned reward function.}
    \label{fig:convnet}
\end{center}
\end{figure}

We implemented our methods on top of a standard open-source implementation of the soft actor-critic algorithm~\cite{haarnoja2018sacapps}. We use standard hyperparameter values used for continuous control problems in this implementation, details of which can be found in Appendix~\ref{app:hyperparams}. The policy and the critic for each task are represented using convolutional neural networks, shown in Figure~\ref{fig:convnet}. It consists of two convolutional layers, each of which is followed by a max-pooling layer, with 8 filters in each of the convolutional layers for simulated tasks, and 32 filters per layer for real world tasks. The flattened output of the convolutional layers is followed by two fully-connected hidden layers with 256 units each. The ReLU non-linearity is applied after each of the convolutional and fully-connected layers. The reward function $f_\psi(\obs)$ is also represented using a convolutional neural network with the same architecture.

We use log-probabilities from a neural network-based classifier as reward for reinforcement learning. However, neural networks are known to drastically change their outputs even with small changes to the input~\cite{Szegedy13}. Thus, they provide a hard decision boundary between different classes, which in our case is similar to running RL with sparse rewards. On the other hand, if the output probabilities of the classifier smoothly transition between positive and negative labels, this would provide a more shaped reward, increasing both the stability and efficiency of the reinforcement learning process. To this end, we found \textit{mixup} regularization to be particularly well-suited for smoothing the classifier predictions~\cite{mixup}. 
We briefly summarize this technique here. Let $\mathbf{s}_i, \mathbf{s}_j$ be any two inputs to the classifier---either from the replay buffer, or from the set of human-provided goal examples---and let $y_i,y_j$ be the corresponding labels. Mixup regularization takes these input/output pairs, and generates the following \textit{virtual} training distribution:
\begin{equation}
\label{eq:mixup}
  \begin{array}{l}
  \Tilde{\mathbf{s}} = \lambda \mathbf{s}_i + (1 - \lambda) \mathbf{s}_j\\
  \Tilde{y} = \lambda y_i + (1 - \lambda) y_j,
  \end{array}
\end{equation}
where $\lambda \sim \text{Beta}(\alpha, \alpha)$. The mixup hyperparameter $\alpha$ controls the extent of mixup, and higher $\alpha$ corresponds to a higher level of mixup (i.e., the sampled $\lambda$ are closer to 0.5 than to 0). \citet{mixup} showed that mixup enables smoother transitions between different classes by encouraging linear behavior, and our experiments indicate that it indeed makes the learned reward function smoother and more amenable to reinforcement learning. For details, see Appendix~\ref{app:hyperparams}.

\section{Simulated Experiments}

\begin{figure}
\begin{center}
    \includegraphics[width=0.8\linewidth]{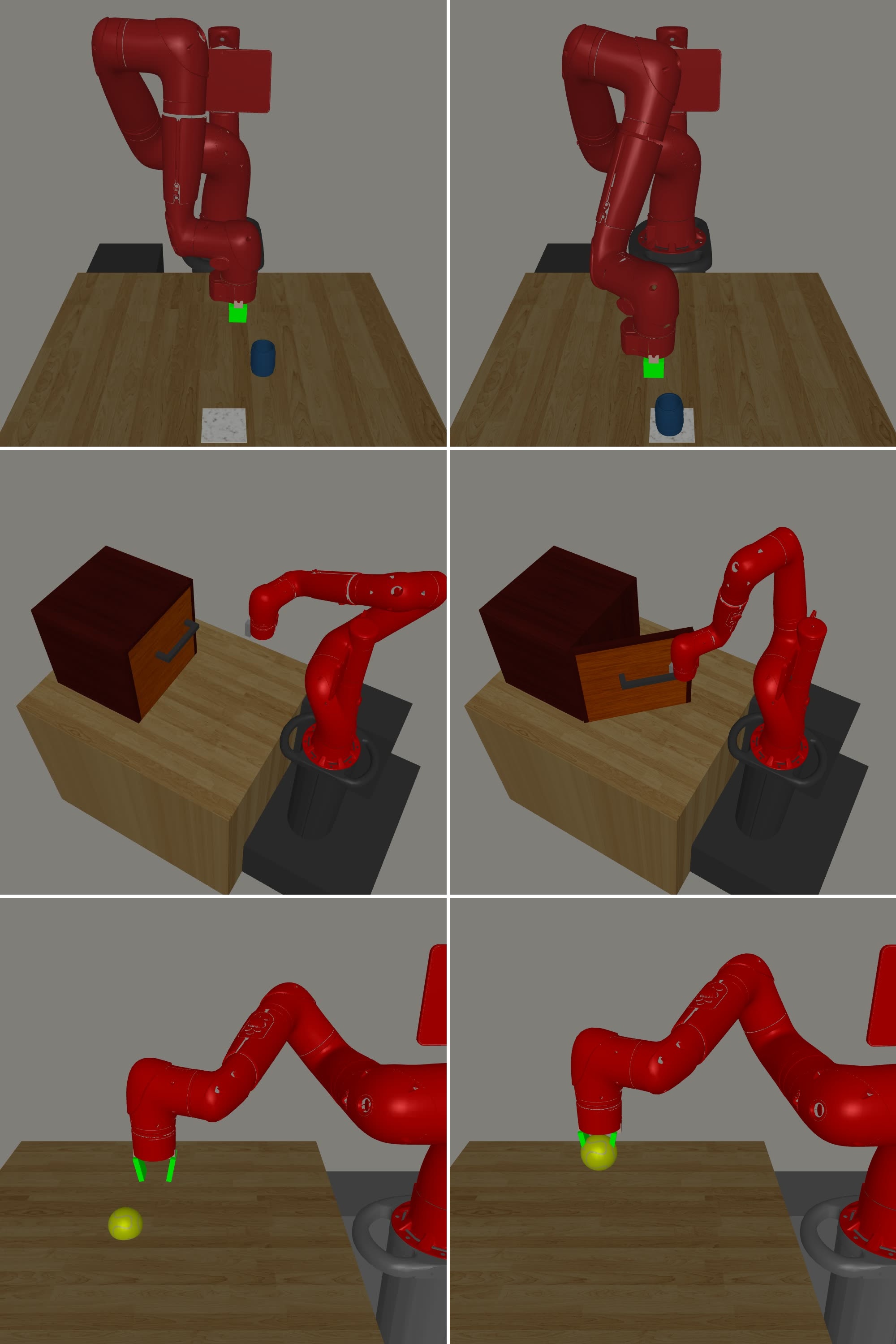}
    \caption{\textbf{Simulated tasks.} The left and right columns depict possible starting states and goal states for each task. In the \textit{Visual Pusher} task (top), the goal is to push a mug onto a coaster, with a randomized initial position of the mug. The middle row shows the \textit{Visual Door Opening} task, where the goal is to open a door of a cabinet by 45 degrees. Initially, the door is either completely closed with probability 0.5, or open up to 15 degrees. In the \textit{Visual Picker} task (bottom), the goal is to pick up a tennis ball from a table and hold it at a particular spot 20cm above the table. The initial position of the tennis ball on the table is randomized. 
    }
    \label{fig:sim_tasks}
\end{center}
\end{figure}

\begin{figure}
\begin{center}
    %\includesvg[width=0.95\linewidth]{figures/sim_all_seeds}
    \includegraphics[width=0.95\linewidth]{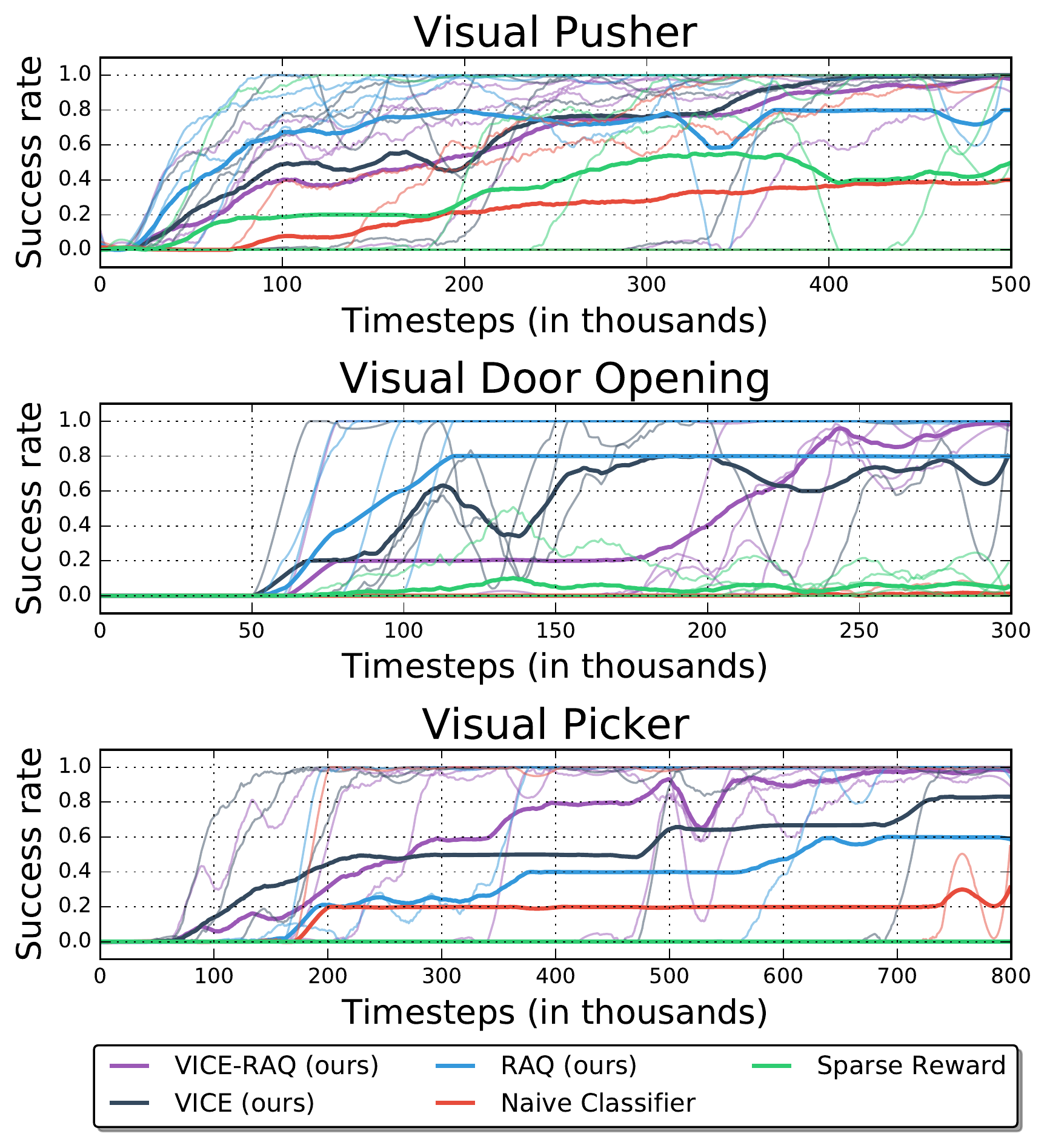}

    \caption{Results on simulated tasks. Each method is run with five different random seeds for each task. The lines in bold indicate the mean across five runs, while the faint lines depict the individual random seeds for each method. We observe that VICE-RAQ achieve the best performance on all tasks, with RAQ being comparable to VICE-RAQ on the Visual Pusher task. We also notice that both RAQ and VICE have significant variance among runs, while VICE-RAQ achieves relatively low variance towards the end of the learning process.}
    \label{fig:sim_results}
\end{center}
\end{figure}

Our simulated experiments are aimed at providing a rigorous comparison between RAQ (Section~\ref{sec:raq_description}), our off-policy extension of VICE (Section~\ref{sec:off-policy-vice}), VICE-RAQ (Section~\ref{sec:viceraq_description}), and standard classifier-based rewards~\cite{flo, Vecerik18} (summarised in Algorithm \ref{alg:classifierRL}), as well as a comparison to RL with sparse rewards. The ability to run multiple trials for every method on multiple tasks allows us to provide a detailed comparison, and we have made an open source release of our code to reproduce all experiments in this section\footnote{\url{https://github.com/avisingh599/reward-learning-rl}}.
The simulated experiments are conducted on three different tasks, illustrated in Figure~\ref{fig:sim_tasks}.
Each of the tasks is performed using end-effector position control with a 7-DoF robotic arm modeled on the Rethink Sawyer. For the picking task, the policy can also continuously control the opening and closing of the gripper. The observation space of the robot is a 48x48 RGB image. No other observations (such as joint angles or end-effector position) are provided to the policy. The goal is specified using a set of goal images, as depicted in Figure~\ref{fig:sim_tasks}. 
We start with 10 goal examples for each of the task, and perform an active query once every 25K timesteps for the \textit{Visual Pusher} task, once every 10K timesteps for \textit{Visual Door Opening} task, and once every 1K timesteps for the \textit{Visual Picker} task. More details about the tasks can be found in Appendix~\ref{app:sim_tasks}.

We perform runs with five different random seeds for every method and task. The results of our experiments are shown in Figure~\ref{fig:sim_results}, and videos can be found on our project website\footnote{\url{https://sites.google.com/view/reward-learning-rl/home}}.
Separate learning curves for all random seeds for each method can be found in Appendix~\ref{app:learning curves}. 
For the pushing task, we observe that both off-policy VICE and VICE-RAQ perform well, and solve the task for all random seeds, while RAQ, the na\"ive classifier and the ground truth sparse reward baselines only succeed for some of the runs. For the harder door opening task, we observe that VICE-RAQ outperforms all other methods, and is able to achieve a success rate of nearly 100\% across random seeds. The performance of off-policy VICE and RAQ is comparable for this task, with some runs failing and others succeeding for both the methods. The nai\"ive classifier and the sparse reward baselines completely fail for all seeds but one. We see a similar pattern for the \textit{Visual Picker} task, with VICE-RAQ strongly outperforming all other methods, and na\"ive classifier-based reward and sparse rewards failing to solve the task for most seeds. Our off-policy extension of VICE slightly outperforms RAQ for this task.

\section{Real-World Experiments}
\label{sec:experiments_real}

\begin{figure}
\begin{center}
    \includegraphics[width=0.8\linewidth]{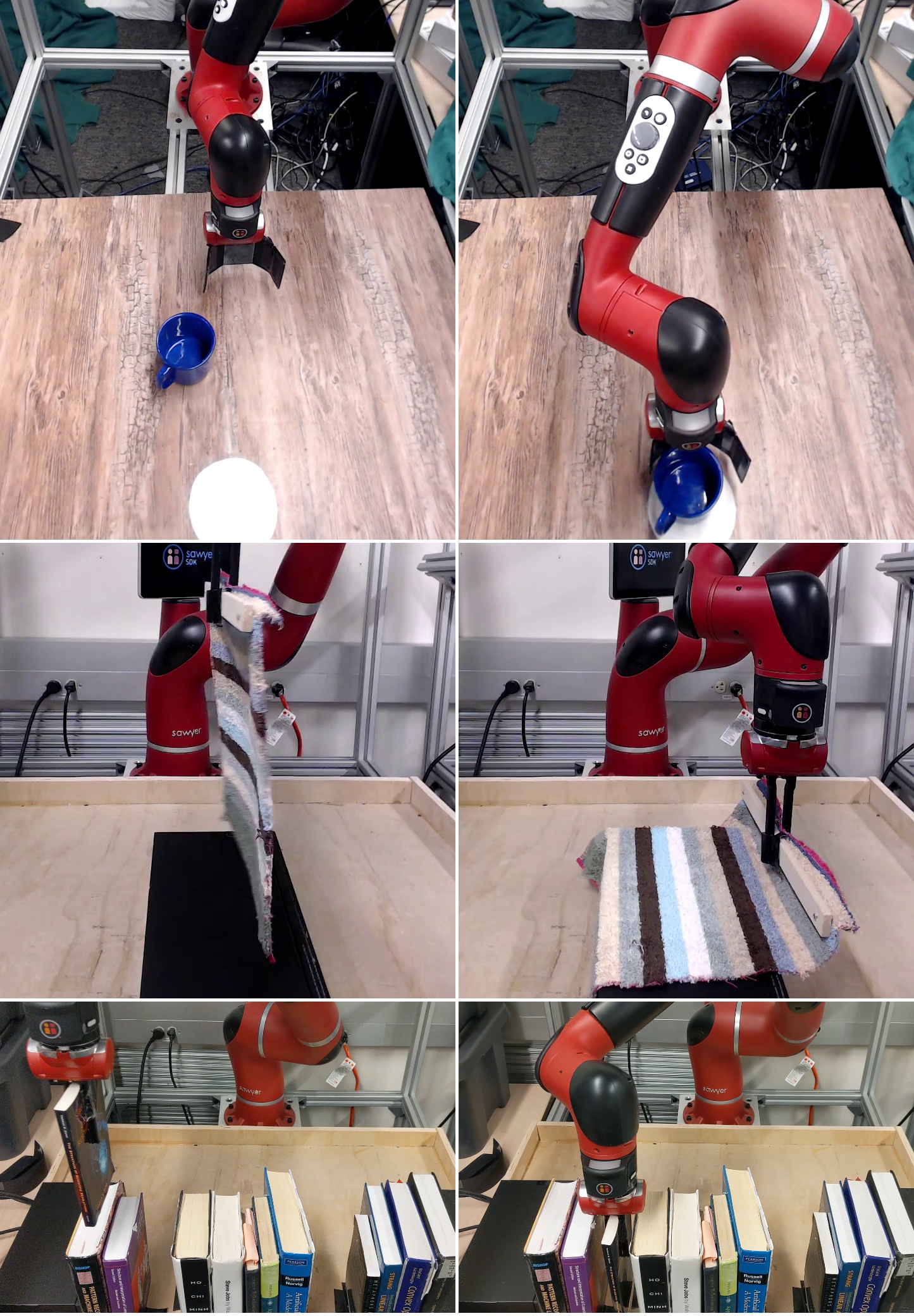}
    \caption{\textbf{Real-world tasks.} The left column depicts possible starting states of the task, while the right column depicts possible goal states. The top row shows the \textit{Visual Pusher} task, in which the goal is to push a mug onto a coaster, and the initial position of the mug is randomized. The middle row presents the \textit{Visual Draping} task, where the goal is to drape a cloth over an object. The bottom row depicts the \textit{Visual Bookshelf} task, where the goal is to inset a book in one of the multiple empty slots in the bookshelf. 
    }
    \label{fig:real_tasks}
    %\vspace{-0.1cm}
\end{center}
\end{figure}

Our real world experiments aim to study end-to-end reinforcement learning from pixels without manually engineered rewards nor instrumentation of the environment to measure rewards.
We evaluate all of our methods, \methodName{}, off-policy VICE, and VICE-\methodName{}, on three complex tasks from vision: pushing a mug onto a coaster, draping a cloth over a box, and a task that requires the robot to insert a book onto a shelf between other books. We also provide a na\"ive classifier-based baseline (summarised in Algorithm \ref{alg:classifierRL}) as a comparison for all of the tasks. The goal of these experiments is to verify that VICE-\methodName{} can successfully learn complex tasks, including non-prehensile manipulation (pushing),  tasks with multiple success conditions (where the book can be placed in one of several locations), and tasks with deformable objects (cloth draping). For all our experiments, we use end-effector position control on a 7 DoF Sawyer robotic manipulator, and our observations consist of an RGB image of size 84 $\times$ 84. We do not make use of robot joint angles or end effector-positions. The final success rates of the trained policies for each of these tasks are shown in Figure~\ref{fig:robot_results}. We first discuss the individual tasks, and then the experimental results.

\begin{figure}
\begin{tabular}{l|c c c c}
            &  VICE-\methodName{} &  \methodName{}  &  VICE  & Na\"ive  \\
            & \small (ours) & \small (ours) & \small (ours) & Classifier\\
 \hline
%  reaching    & 3.6   &  1.0  & 4.1     & 12.3  \\
 \!\!\!visual pushing     & 100\%   &  60\%  & 0\%    & 0\% \\
 \!\!\!visual draping     & 100\%   &  100\%  & 100\%    & 0\% \\
 \!\!\!visual bookshelf \!\!\! & 100\% &  0\%  & 60\%  & 0\%

\end{tabular}

\caption{Results on the real world robot experiments. For all tasks, the reported numbers are success rates, indicating whether or not the object was successfully pushed to the goal, whether the cloth was successfully draped over the able, and whether the book was placed correctly on the shelf, averaged across 10 trials. In all cases, VICE-\methodName{} succeeds at learning the task, while VICE and \methodName{} succeed at some tasks while failing at others.}
\label{fig:robot_results}
\end{figure}

\paragraph{Visual Pushing} This non-prehensile manipulation task (depicted in Figure~\ref{fig:real_tasks}) requires the robot to push a mug onto a coaster. The position of the mug and coaster must be inferred from the images, and the initial position of the mug varies between different trials. In order to succeed, the robot should push the mug such that is gets placed completely within the coaster. Here, the robot must make use of the images to determine whether it has achieved the goal successfully, and the increased challenge of non-prehensile manipulation allows to better differentiate the performance of the different methods.

\begin{figure}
\begin{center}
    \includegraphics[width=0.80\linewidth]{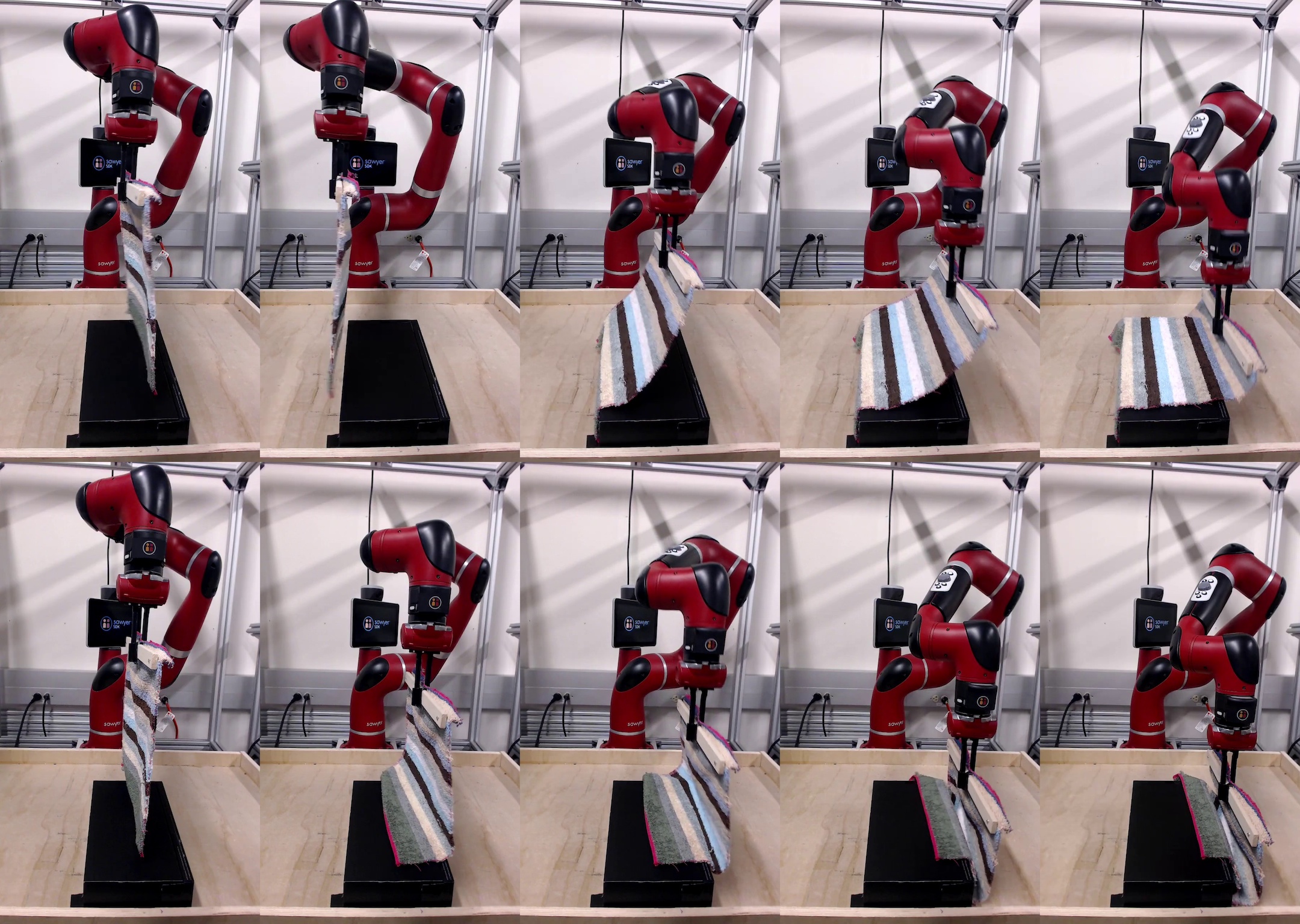}
    \caption{In this figure, we demonstrate why learning a reward function on pixels is necessary for solving complex tasks in the real world. The task here is to drape a cloth over a box. The top row shows a rollout from the final policy trained by our method, while the bottom row shows a rollout from a policy trained on a hand-defined reward on robot state alone. Our policy is able to successfully drape the cloth over the box, while the policy trained without vision only sees the end-effector position, which it succeeds in moving to the right place, but fails to drape the cloth on the box.}
    
    \label{fig:drape_comparison}
\vspace{-0.8cm}
\end{center}
\end{figure}

\paragraph{Visual Draping} This task requires draping a cloth over a box -- essentially a miniaturized version of a tablecloth draping task. This task is depicted in Figures~\ref{fig:real_tasks} and~\ref{fig:drape_comparison}. The robot starts with holding the cloth in its gripper over the box. In order to succeed, it must drape the cloth smoothly, without crumpling it and without creating any wrinkles. In order to demonstrate the challenges associated with this task, we ran a baseline that only used the robot's end effector position as observation and a hand-defined a reward function on this observation (Euclidean distance to goal). We observed that this baseline failed to achieve the objective of this task, as it simply moved the end effector in a straight line motion to the goal, while this task cannot be solved using any straight line trajectory. See Figure~\ref{fig:drape_comparison} for more details. 

\begin{figure}
\begin{center}
    \includegraphics[width=\linewidth]{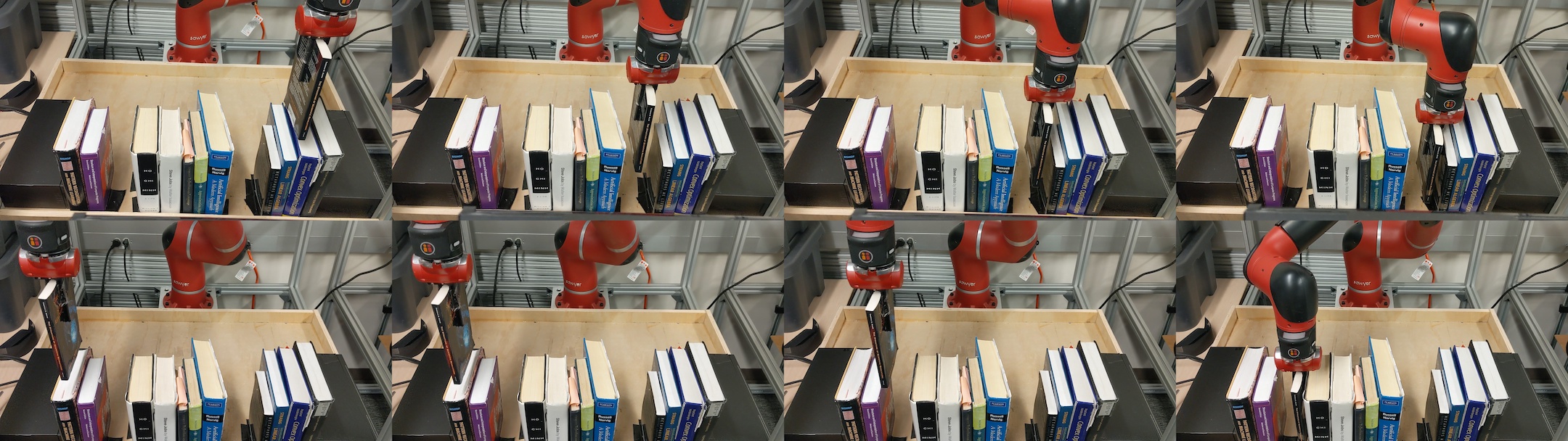}
    \vspace{-0.4cm}
    \caption{In this figure, we demonstrate how classifiers are more expressive than goal images for describing a task. The goal for this task is place a book in any empty slot in a bookshelf, and the initial position of the robot arm holding the book is randomized. The top row shows a rollout when the book starts on the right, while the bottom row shows a rollout when the book starts on the left. Here, we see that our method learns a policy to insert the book in different slots in the bookshelf depending on where the book is at the start of a trajectory. The robot usually prefers to put the book in the nearest slot, since this maximizes the reward that it can obtain from the classifier. On the other hand, if we were using goal images to specify the task, the robot would always place the book in one of the two slots, regardless of the starting position of the book. }
    \label{fig:book_multigoal}
\end{center}
\vspace{-0.8cm}

\end{figure}

\paragraph{Visual Bookshelf} In this task, the goal is to insert a book into an empty slot on a bookshelf. The task is depicted in Figures~\ref{fig:real_tasks} and~\ref{fig:book_multigoal}. The initial position of the arm holding the book is randomized, requiring the robot to succeed from any starting position. Crucially, the bookshelf has several open slots, which means that, from different starting positions, different slots may be preferred. We chose this task to emphasize that a goal classifier is fundamentally different from a goal state: there is not a single ``goal image'' that represents success at the task, but rather a condition on the position of the book that can be fulfilled in different ways (see Figure~\ref{fig:book_multigoal}). The successful outcome examples correspondingly illustrate successful placements in both positions. 

We provide 80 success examples each for the pushing, draping and book placing tasks. We query once every 250 timesteps for the pushing and book placing task, while querying once every 500 timesteps for the draping task. The \textit{Visual Pusher} experiments are run for 6.2K timesteps (about 90 minutes of real world time), the \textit{Visual Draping} experiments are run for 25K timesteps (about 4 hours), and the \textit{Visual Bookshelf} experiment is run for 19K timesteps (about 3 hours). We therefore make 25 queries for the pushing experiment, 50 queries for the draping experiment, and 75 queries for the book placing experiment. Note that all of these tasks are learned directly from raw images, making these training times very efficient as compared to prior methods, including methods that use ground truth rewards~\cite{qt-opt, haarnoja2018sacapps, Levine16googlegrasping, pinto2016supersizing}. We hypothesize that the regularized classifiers learned by our method provide favorable reward shaping that makes image-based RL not only more practical, by not requiring engineered rewards, but also substantially more efficient.

The results of our experiments are provided in Figure~\ref{fig:robot_results}, and videos of the tasks are provided on the project website\footnote{\url{https://sites.google.com/view/reward-learning-rl/home}}. All the tasks are evaluated in terms of success rates.

The \textit{Visual Pushing} task requires the robot to interpret the RGB camera images and deal with variability in the mug placement. For this task, VICE-\methodName{} obtains a success rate of 100\%, while  \methodName{} only obtains a success rate of 60\%. Both off-policy VICE and na\"ive classifier fail to solve this task. This indicates that including active queries in the classifier training process is helpful for obtaining good rewards, both with and without VICE. These experiments further show that VICE-RAQ can improve upon RAQ via leveraging the data collected by the policy during the reinforcement learning process. 

For the \textit{Visual Draping} task, we observe that all of our reward-learning methods (off-policy VICE, VICE-RAQ and RAQ) are able to solve the task, and only the na\"ive classifier baseline fails. We also compare to a baseline that is based on a reward defined on the robot state alone, in this case, the 3D position of the end-effector (the reward being the Euclidean distance to a goal end-effector position). We observe that this baseline fails to solve the task (see Figure~\ref{fig:drape_comparison}). This indicates that performing this task requires the robot to actually pay attention, visually, to the deformations in the cloth, in order to perform the draping successfully. Manually designing reward functions for such deformable object manipulation tasks is generally extremely difficult, but all variants of our method are able to handle it successfully.

For the \textit{Visual Bookshelf} task, where success corresponds to whether the book was successfully placed in an empty slot on the bookshelf or not), we see that \methodName{} alone is unable to solve this task, while off-policy VICE learns a policy that only succeeds sporadically. The policy learned with VICE-\methodName{} solves this task consistently, indicating that the combination of query labels and negative labels from all visited states from VICE provide improved classifier training.

\section{Discussion and Future Work}

In this paper, we proposed an approach to reinforcement learning without hand-programmed reward functions. Our method, which we call VICE-RAQ, constructs a reward function from a modest number of user-provided examples of successful outcomes, which in practice might consist simply of pictures of the scene where the task has been successfully completed. Such examples are often substantially easier for a user to provide than either hand-programmed reward functions or full demonstrations. The initial reward is constructed out of a classifier trained on these examples and adversarially mined negatives. Beyond the initially provided success examples, our method uses a modest number of active queries, where the user is asked to label outcomes achieved by the robot as either successful or not. These additional queries are also simple to provide, and roughly correspond to the user directly reinforcing the robot's behavior. However, the user does not need to label all of the robot's experience -- only about 50 queries are used in our experiments, out of tens of thousands of transitions.

While our method improves substantially on both na\"{i}ve classifier rewards and VICE in our experiments, it does have a number of limitations. First, the requirement to obtain labels from the user imposes additional assumptions. Second, the number of queries required---around 50 per training run---is still non-trivial. It is quite possible that a more intelligent query criterion could reduce this number further, and a promising direction for future work is to incorporate techniques for quantifying model uncertainty in the classifier, such as Bayesian neural networks. Lastly, our method does not benefit from any shared structure \emph{between} tasks. In reality, tasks have considerable shared structure, and a classifier that incorporates data from multiple tasks, analogously to prior work on meta-learning~\cite{flo,xu2018learning}, could in principle further reduce the number of queries and examples that is needed.

By enabling robotic reinforcement learning without user-programmed reward functions or demonstrations, we believe that our approach represents a significant step towards making reinforcement learning a practical, automated, and readily usable tool for enabling versatile and capable robotic manipulation. By making it possible for robots to improve their skills directly in real-world environments, without any instrumentation or manual reward design, we believe that our method also represents a step toward enabling lifelong learning for robotic systems that learn directly ``in the wild.'' This capability can make it feasible in the future for robots to acquire broad and highly generalizable skill repertoires directly through interaction with the real world.

%% Use plainnat to work nicely with natbib. 

\section*{Acknowledgements}
This research was supported by the National Science Foundation via IIS-1651843 and IIS-1700696, the Office of Naval Research, Google, NVIDIA, and Amazon. The authors would like to thank Michael Chang, Oleh Rybkin and Aviral Kumar for their feedback on early drafts of this paper, and May Simpson for her help with choosing a background color for simulated robot environments. The authors would also like to thank Shikhar Bahl and Sudeep Dasari for their assistance with the robot experiments, and all the students and post-docs at RAIL for support and feedback.

\bibliographystyle{plainnat}
\bibliography{references}

%\newpage
\clearpage
\onecolumn

\appendix
\section{Experimental details}
\subsection{Simulated tasks}
\label{app:sim_tasks}
Detailed descriptions of our simulated tasks are provided here.

\paragraph{Visual Pusher}
The goal of this task is to push a mug onto a coaster. The robot end effector is constrained to move in a 2D XY plane, and the initial position of the mug is randomized over a 20cm x 15cm region. A success is when the final position of the mug is within 3cm of the goal. 
\paragraph{Visual Door}
The goal of this task is to open a door by 45 degrees. Initially, the door is either completely closed (with probability 0.5), or open up to an angle of 15 degrees. The robot end effector is equipped with a hook, and the robot needs needs to trap the handle of the door in its hook and pull it open. The robot end effector is allowed to move in the full 3D XYZ space.
\paragraph{Visual Picker}
The goal of this task is to pick up a tennis ball from a table. The robot end effector is free to move in the 3D XYZ space, and the initial position of the ball is randomized over a 10cm x 10cm square region over the table. Along with the robot end-effector, we also control the opening and closing of a parallel jaw gripper. The robot needs to pick up the ball and move it to a fixed location that is 20 cm above the table. A success is when the final position of the ball is within 3cm of the goal.

\subsection{Hyperparameter details}
\label{app:hyperparams}
\paragraph{Parameters for Soft Actor-Critic} We did not tune any hyperparameter of the soft actor-critic algorithm, and used default values provided in the authors' open-source implementation~\cite{haarnoja2018sacapps}. These value are a learning rate of 3e-4 on the Adam optimizer~\cite{adam}, a batch size of 256, $\tau=0.05$, a discount factor of 0.99, a target update frequency of 1, and a target entropy of $\frac{1}{d_a}$, where $d_a$ is the action dimension for the environment.

\paragraph{Parameters for Discriminator Training} We train the discriminator also using Adam with a learning rate of 3e-4 and a batch size of 256. We take $N$ update steps on the discriminator at the end of every epoch of RL updates (where each epoch consists of 1000 timesteps). We sweeped over the value of $N$ individually for each of our methods and all the simulated tasks. We sweeped over N=[5,10,100] and found that N=10 works well for all task/method combinations, except the Visual Door Opening task with VICE-RAQ, where N=5 performs slightly better. We also sweeped over using mixup (with parameter $\alpha=1$) and not using mixup, and found that mixup greatly helped for the Visual Door Opening and the Visual Picker task, but slightly decreased the performance for the Visual Pushing task. After the hyperparameter sweep, fresh runs were made for all tasks and methods for reporting the final results.

\subsection{Detailed learning curves}
\label{app:learning curves}

\begin{figure*}[h]
\begin{center}

    % \includesvg[width=0.95\linewidth]{figures/sim_all_methods_all_seeds}
    \includegraphics[width=0.95\linewidth]{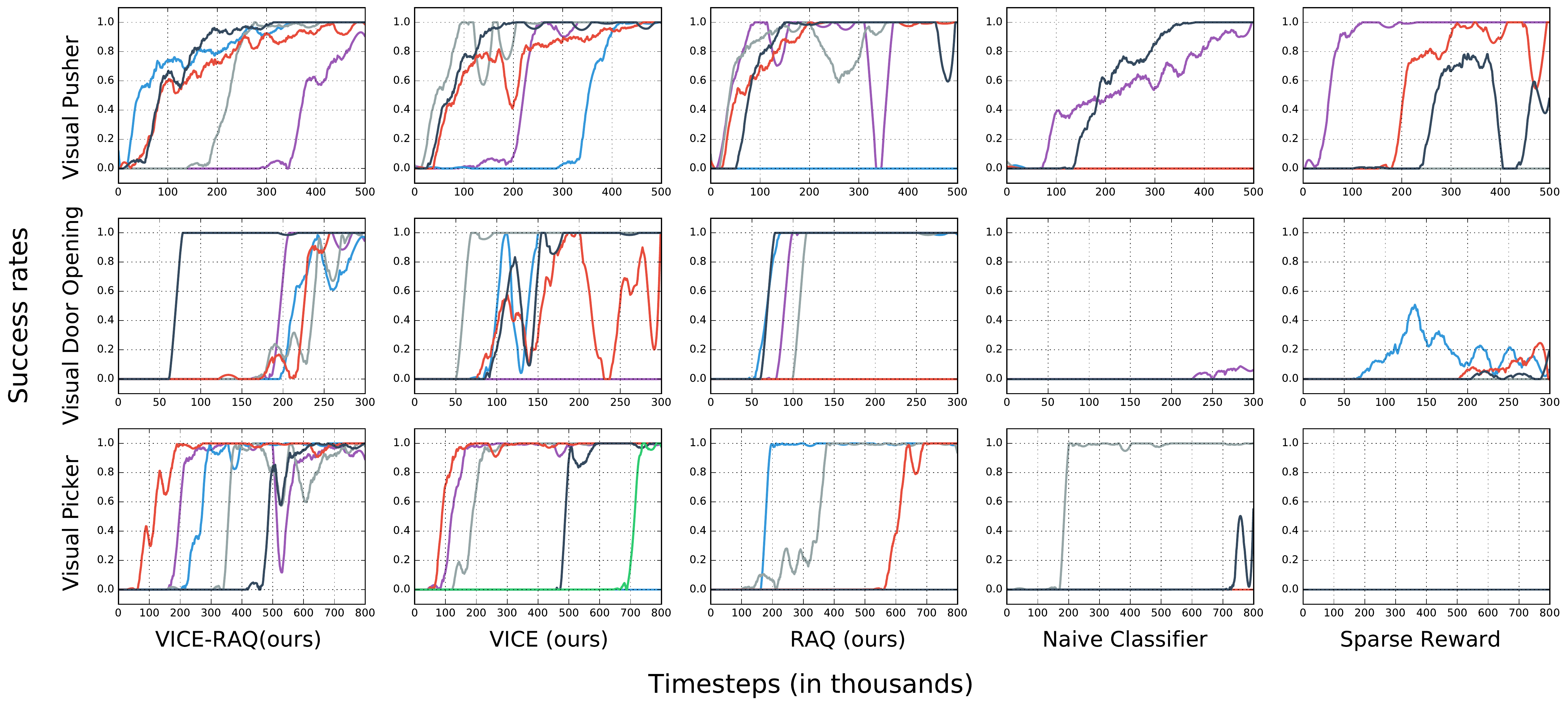}
    \label{fig:sim_all_methods_all_seeds}
    \caption{This plot shows the results on all of our simulated tasks for all of our methods and baselines. Each plots shows results from each of the five random seeds run for that task and method.}
    %\vspace{-0.1cm}
\end{center}
\end{figure*}

\begin{figure*}[h]
\begin{center}

    \subfigure{\includegraphics[width=0.9\linewidth]{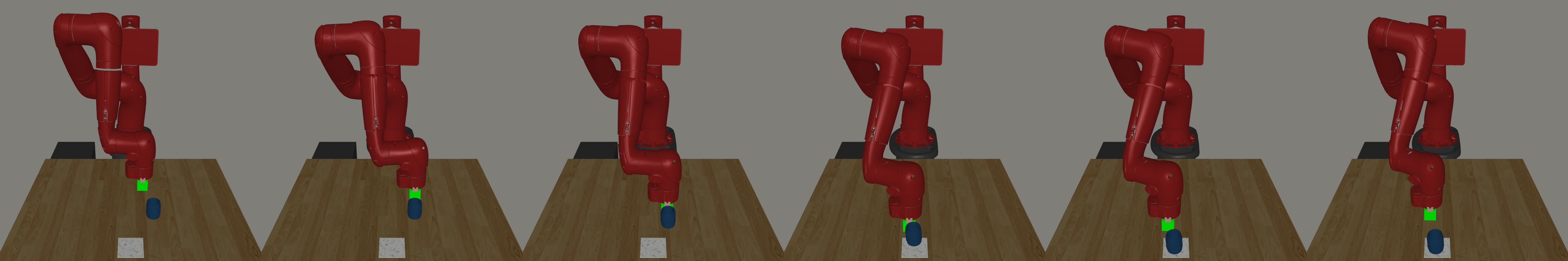}}
    \subfigure{\includegraphics[width=0.9\linewidth]{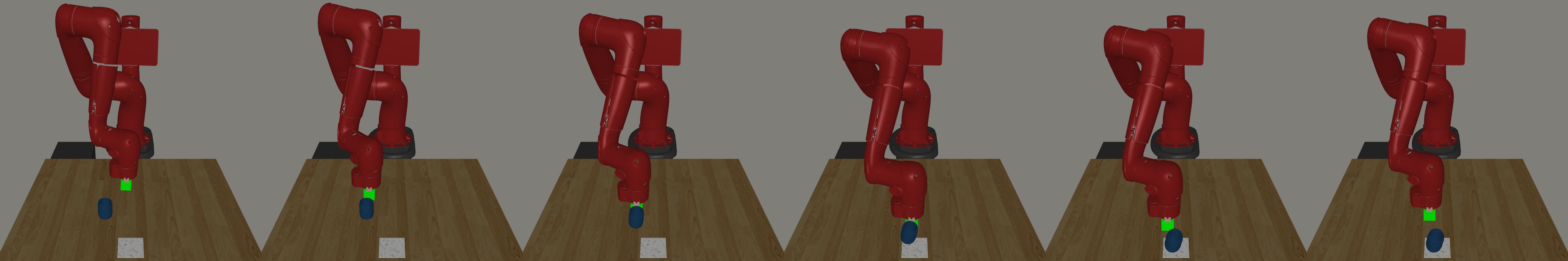}}
    \subfigure{\includegraphics[width=0.9\linewidth]{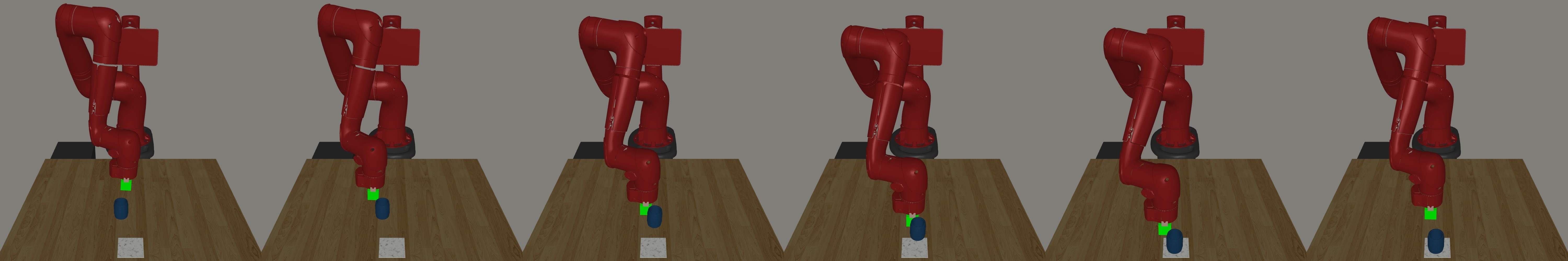}}
    %\includegraphics[width=0.9\linewidth]{figures/push_rollout_01.jpg}
    %\subfigure{\includegraphics[width=0.9\linewidth]{figures/push_rollout_02.jpg}}
    \caption{Example evaluation rollouts for the simulated \textit{Visual Pusher} task from a policy learned using VICE-RAQ.}
    \label{fig:sim_push_rollouts_viceraq}
    %\vspace{-0.1cm}
\end{center}
\end{figure*}

\begin{figure*}[h]
\begin{center}
    \subfigure{\includegraphics[width=0.9\linewidth]{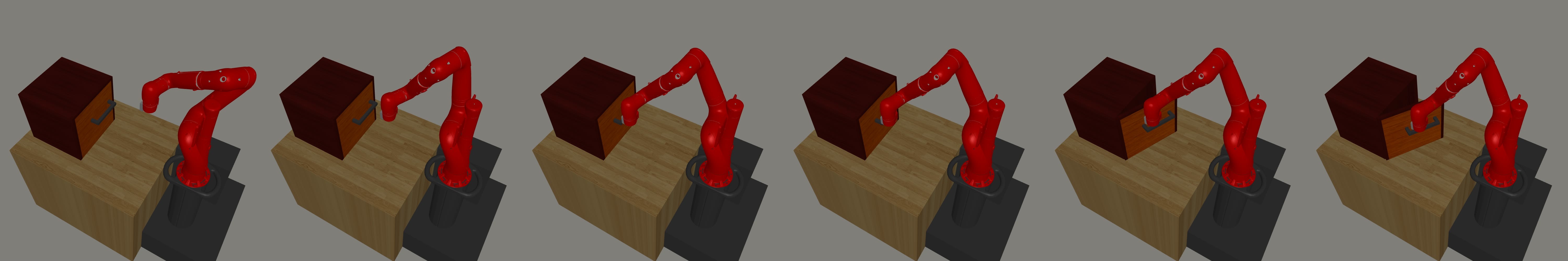}}
    \subfigure{\includegraphics[width=0.9\linewidth]{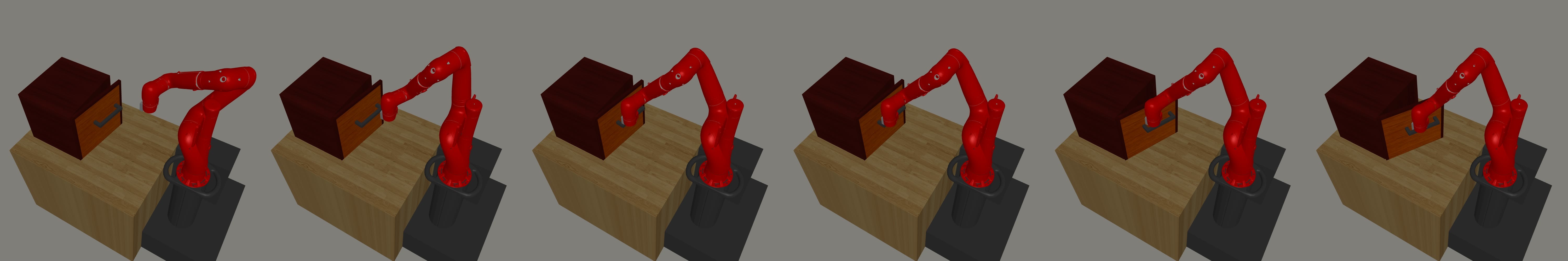}}
    \subfigure{\includegraphics[width=0.9\linewidth]{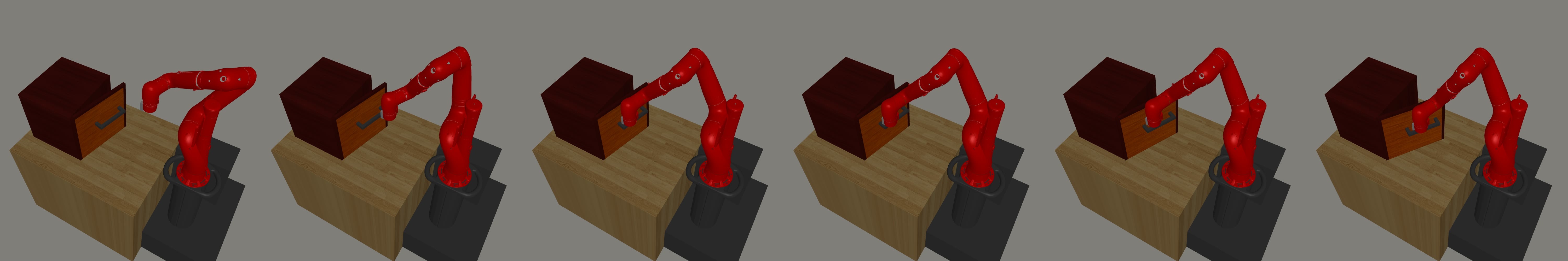}}

    \caption{Example evaluation rollouts for the simulated \textit{Visual Door Opening} task from a policy learned using VICE-RAQ.}
    \label{fig:sim_push_rollouts_viceraq}
    %\vspace{-0.1cm}
\end{center}
\end{figure*}

\begin{figure*}[h]
\begin{center}
    \subfigure{\includegraphics[width=0.9\linewidth]{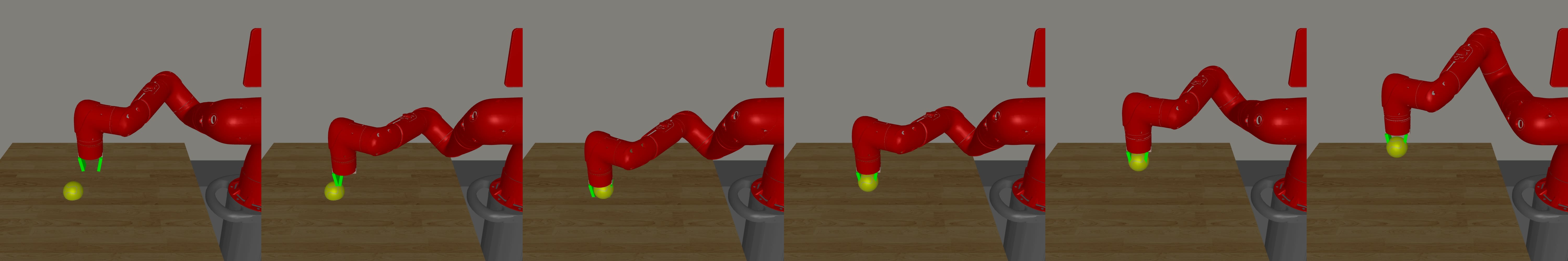}}
    %\subfigure{\includegraphics[width=0.9\linewidth]{figures/pick_rollout_1.jpg}}
    \subfigure{\includegraphics[width=0.9\linewidth]{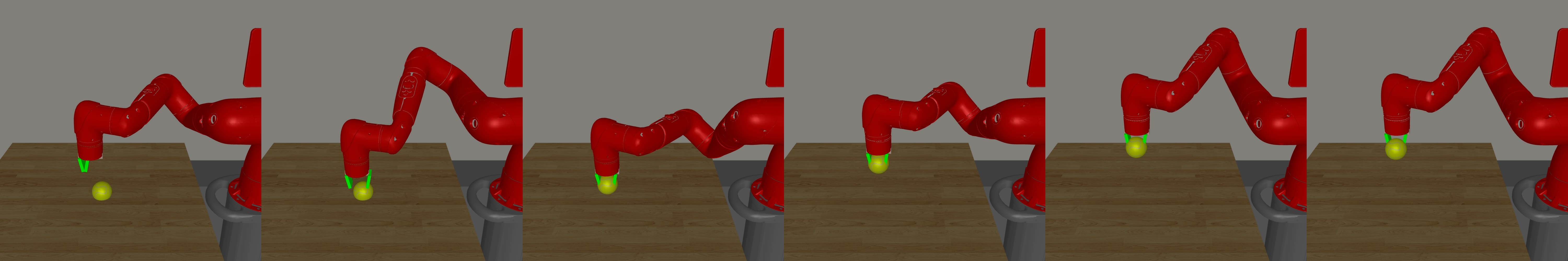}}
    \subfigure{\includegraphics[width=0.9\linewidth]{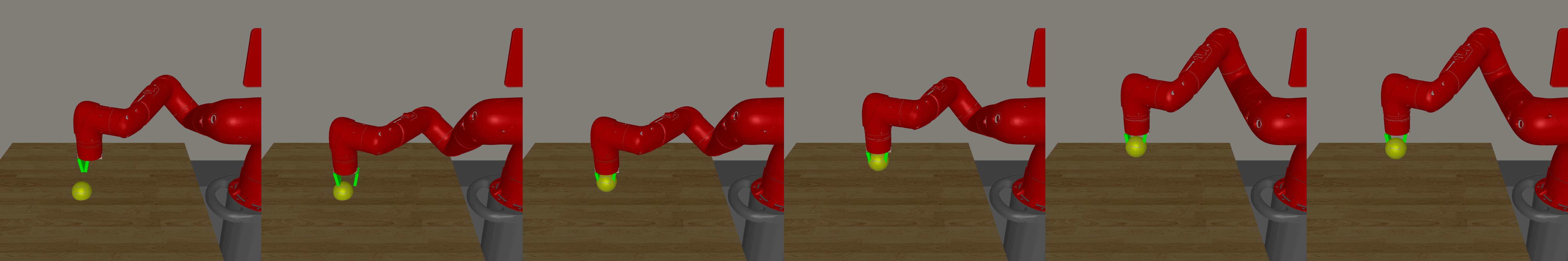}}

    \caption{Example evaluation rollouts for the simulated \textit{Visual Picker} task from a policy learned using VICE-RAQ.}
    \label{fig:sim_push_rollouts_viceraq}
    %\vspace{-0.1cm}
\end{center}
\end{figure*}

\end{document}